\pdfoutput=1

\documentclass[11pt]{article}

\usepackage{acl} 

\usepackage{times}
\usepackage{latexsym}

\usepackage[T1]{fontenc}

\usepackage[utf8]{inputenc}

\usepackage{microtype}

\usepackage{graphicx}
\usepackage{epstopdf}
\usepackage{multirow}
\usepackage{multicol}
\usepackage{arydshln}
\usepackage{enumitem}
\usepackage{graphicx}
\usepackage{color}
\usepackage{xcolor}
\usepackage{listings}
\title{Evidence-based Interpretable Open-domain Fact-checking \\ with Large Language Models}

\author{Tan Xin \and Zou Bowei \and Aw Ai Ti \\
  Institute for Infocomm Research (I2R), A*STAR \\
  \texttt{\{tan\_xin,zou\_bowei,aaiti\}@i2r.a-star.edu.sg} \\ 
}

\begin{document}
\maketitle
\begin{abstract}
Universal fact-checking systems for real-world claims face significant challenges in gathering valid and sufficient real-time evidence and making reasoned decisions.
In this work, we introduce the \textbf{O}pen-domain \textbf{E}xplainable \textbf{Fact}-checking (OE-Fact) system for claim-checking in real-world scenarios. 
The OE-Fact system can leverage powerful understanding and reasoning capabilities of large language models (LLMs) to validate claims and generate causal explanations for fact-checking decisions.
To adapt the traditional three-module fact-checking framework to the open domain setting, we first retrieve claim-related information as relevant evidence from open websites.
After that, we retain the evidence relevant to the claim through LLM and similarity calculation for subsequent verification.
We evaluate the performance of our adapted three-module OE-Fact system on the Fact Extraction and Verification (FEVER) dataset.
Experimental results show that our OE-Fact system outperforms general fact-checking baseline systems in both closed- and open-domain scenarios, ensuring stable and accurate verdicts while providing concise and convincing real-time explanations for fact-checking decisions.
\end{abstract}

\section{Introduction}

In an era of rapid proliferation of non-factual data, inaccurate and misleading content poses significant challenges to humanity, preventing people from making informed decisions and eroding trust in online resources.
The vast spread of misinformation highlights the importance of automatic fact-checking tasks in natural language processing (NLP) and calls for the emergence of robust and effective fact-checking systems.

Existing effective fact-checking models~\citep{nie2019combining,zhong2020reasoning,dehaven2023bevers,soleimani2020bert,jiang2021exploring,pradeep2021scientific,pradeep2021vera} usually adopt a three-module pipeline to verify the authenticity of a given claim. Firstly, the document retrieval module retrieves claim-related documents from a database prepared in advance. 
Secondly, the sentence selection module is harnessed to predict evidence for a claim by scoring and ranking the sentences from retrieved documents.
Lastly, the claim classification module is used to verify the claim's authenticity based on the predicted evidence.
Although the traditional three-module framework effectively improves claim-checking by training three modules in database-centric scenarios, it remains challenging in open-domain real-world scenario claim-checking, which requires more valid and sufficient real-time evidence.

Unlike database-centric scenarios, open-domain fact-checking tasks are more challenging due to the complexity and continuous evolution of real-world claims.
In this situation, the claim-checking system needs to first understand the claims entered by users in informal language, then retrieve evolving real-time claim-related evidence from multiple sources, and then select claim-relevant evidence from large amounts of candidate evidence to reach reliable and reasoned verdicts.

To achieve such a fact-checking system that focuses on real-world claims, we propose an \textbf{O}pen-domain \textbf{E}xplainable \textbf{F}act-checking system (OE-Fact). 
To cater to the nature of real-world claim-checking, we take advantage of the understanding and reasoning capabilities of large language models (LLMs) to build an LLM-based fact-checking system that generates reliable judgments and reasonable explanations for a given claim.
Specifically, we adopt the traditional three-module pipeline in OE-Fact and adjust the three modules to adapt to the open-domain scenario: 
First, we employ an evidence retrieval module to retrieve claim-related information from open websites as candidate evidence. 
Afterward, we deploy an evidence selection module to filter the evidence most related to the claim (top-5) through LLM and semantic similarity. 
Finally, a verdict generation module is designed to generate predictive labels and real-time decision explanations of claims based on preserved claim-relevant evidence.

The main contribution of this work is three-fold: 
\begin{itemize}
    \item We contribute the fact-checking system OE-Fact, which fills the gap in real-world claim-checking within the open-domain setting.
    \item The experimental results on Fact Extraction and VERification (FEVER) dataset~\citep{thorne2018fever} highlight the effectiveness of LLMs in generating stable and accurate judgments in fact-checking tasks.
    \item The real-time fact-checking decision explanation generated by our LLM-based OE-Fact system guarantees the verdict's transparency and enhances the explanation's overall coherence and plausibility.
\end{itemize}

\section{Related Work}

\subsection{Traditional Fact-checking}

Fact-checking is an important task in the field of NLP that helps humans automatically verify the authenticity of claims based on evidence.
Conventional fact-checking works usually consist of three modules, which are used to retrieve documents, sort the evidence in each document, and then make authenticity judgments based on top-k predicted evidence.
Previous works~\citep{nie2019combining,zhong2020reasoning,dehaven2023bevers,soleimani2020bert,jiang2021exploring,pradeep2021scientific,pradeep2021vera} mainly focus on database-centric scenarios that verify claims based on the evidence retrieved from the relevant database prepared in advance. 
In this database-centric closed-domain setting, most required information of a claim is from a single source, e.g., 87\% of claims in the FEVER dataset~\citep{thorne2018fever} only require information from a single Wikipedia article~\citep{ jiang2020hover}.

\subsection{Real-world Open-domain Fact-checking}
\label{section: Real-World}

Unlike general fact-checking tasks, real-world fact-checking tasks are designed to process real-world user-input claims related to news and trends based on real-time evidence retrieved from open websites.
Compared with the traditional database-centric fact-checking task, the challenges of fact-checking in the real-world open-domain setting are exacerbated by the following key factors:

\begin{itemize}[leftmargin=*]
    \item \textbf{Complexity:} Both user-input claims and evidence information from the Internet are often informal, which may contain semantic ambiguities and complex syntactic structures requiring advanced language understanding capabilities.
    \item \textbf{Diversity:} Evidence obtained from public websites often comes from multiple sources, resulting in uncertainty, diversity, and even contradictions, which need to be considered without basis to ensure a comprehensive and accurate assessment.
    \item \textbf{Timeliness:} The timeliness of evidence is critical in checking fast-moving real-time events and news. Ensuring adequate and timely evidence is critical for real-world open-domain claim checking.
\end{itemize}

As stated before, this work concentrates on the real-world claim-checking scenario. 
In order to mitigate the above challenges, we leverage LLM coupled with comprehensive understanding and reasoning capabilities to generate reliable judgments for given claims. 

\subsection{Reasoning and Explanation Generation}

A brief explanation of how a decision was made makes the verdict more reliable and convincing, especially for real-world claims~\citep{krishna2022proofver,fajcik-etal-2023-claim}.
Various recent works provide post-hoc explanations for model predictions by highlighting relevant parts of the evidence~\citep{popat2017truth,cui2019defend,yang2019xfake,lu2020gcan}, generating justifications based on knowledge graphs~\citep{gad2019exfakt,ahmadi2019explainable}, or generating a summary of the retrieved relevant evidence~\citep{atanasova2020generating,kotonya2020explainable,jolly2022generating}. 

Different from previous works focusing on post-hoc explanation generation, this work aims to generate a real-time decision explanation during the claim verification process by leveraging LLM to analyze the causal relationship between evidence and claims.
In particular, our work harness the reasoning ability of LLM for real-time interpretation to ensure the verdict's transparency and explanation's coherency.

\section{Approach}

\subsection{Open-domain Evidence Retrieval}

One major challenge for real-world claim checking is retrieving sufficient and real-time claim-related evidence to respond to rapidly evolving and changing events (see Section~\ref{section: Real-World}). 
As one of the world's most widely used search engines, the Google search engine contains timely information from multiple sources and prioritizes relevant and accurate information for a given query, meeting our need for real-time evidence retrieval for real-world claim checking.

To obtain sufficient and relevant claim-related information, in this work, we employ two types of queries to retrieve candidate evidence with Google Custom Search Engine API\footnote{\url{https://developers.google.com/custom-search/v1/overview?hl=zh-cn}}.
On the one hand, we rely on the information interpretation and extraction capabilities of the Google search engine to directly submit the entire user-input claim as a query for evidence retrieval.
On the other hand, to expand the search scope to enhance the retrieval of claim-related information, we further extract and submit relevant critical words as queries to the Google search engine.
Specifically, we first apply the \texttt{en\_core\_web\_sm} model\footnote{\url{https://github.com/explosion/spacy-models/releases/tag/de_core_news_sm-3.3.0}} of the SpaCy toolkit~\citep{honnibal2020spacy} to analyze and extract nouns and noun chunks with ``token'' and ``noun\_chunks'' attributes present in the claims.
Then, we input the query that combines the keyword nouns and noun chunks with ``;'' into the Google search engine for broad claim-related candidate evidence retrieval.

This dual query strategy for candidate evidence retrieval ensures the comprehensiveness and accuracy of claim-related evidence.
It is worth noting that we select the top-10 matching results returned by Google API for each search query in this work to reduce the pressure and challenges caused by massive information for subsequent claim verification.

\subsection{Claim-relevant Evidence Selection}

For traditional three-module fact-checking systems, when there is both supporting and refuting evidence, the systems typically select evidence from one single group based on in-built biases and do not distinguish whether evidence is relevant to the claim~\citep{fajcik-etal-2023-claim}.
This issue is much more severe in open-domain scenarios. When evidence comes from multiple information sources, the noise brought by complex and redundant evidence information will lead to uncertainty, diversity, and even contradiction in information, leading to biased judgments.

In order to alleviate the noise caused by complex and redundant evidence retrieved from multiple sources, this work adopts an evidence selection module that sequentially employs Llama and semantic similarity calculation to select claim-relevant evidence to gain reliable verdicts.

\paragraph{LLM-based Evidence Filtering.} Multiple NLP tasks have witnessed the powerful understanding and analysis capabilities of large language models~\citep{touvron2023llama} for human natural language.  
Inspired by this, we propose to harness the foundation model to filter out evidence irrelevant to the claim.
Specifically, we provide claim and candidate evidence as input and employ a 1-shot prompt to instruct LLM to return claim-relevant evidence as follows:

\vspace{2ex}
\begin{tabular}{p{6.5cm}} 
\emph{List the sentences most relevant to \colorbox{lightgray}{Claim} from \colorbox{lightgray}{Evidence} and your own knowledge. Your output is: \textcolor{gray}{Sentences.}}\\
\end{tabular}
\vspace{1.5ex}

\paragraph{Similarity-focused Evidence Selection.} In order to avoid losing claim-relevant information in the LLM-based evidence filtering stage, this stage further utilizes BERT-based similarity calculation to select relevant evidence to contribute robust and reliable information for the claim verification and decision interpretation module.
In detail, for each piece of evidence retained by the LLM-based evidence filtering stage, we first calculate the cosine similarity between it and all candidate evidence based on Bidirectional Encoder Representations from Transformers (BERT)~\citep{kenton2019bert} embeddings.
After calculating the similarity scores of all candidate evidence, we select the top 5 pieces with the highest scores as evidence for further claim verification and explanation generation.

\subsection{Verdict and Real-time Fact-checking Decision Explanation}

Providing fact-checking decision explanations is critical for understanding the verdicts.
Generating real-time verdict explanations during claim verification can improve transparency and reduces potential bias during verification.

To obtain such real-time decision explanations, we verify claims and generate causal explanations for verdicts in a single module.
To achieve this goal, we provide claims and evidence from the evidence selection module as input to LLM and guide it to verify the claim with the label ``True'', ``False'' or ``Uncertain'' and generate real-time decision explanations simultaneously.
In this process, we stimulate the reasoning ability of LLM through a 1-shot prompt, inspiring it to analyze the causal relationship between evidence and claims as follows:

\vspace{3ex}
\begin{tabular}{p{6.5cm}} 
\emph{Verify the truth of \colorbox{lightgray}{Claim} based on \colorbox{lightgray}{Evidence}, with label `True', `False', or `Uncertain', and explain why you get this conclusion. Your output is: \textcolor{gray}{Label}; \textcolor{gray}{Explanation}.} \\
\end{tabular}
\vspace{3ex}

The real-time decision explanations we generate in this module provide timely, transparent explanations that increase the effectiveness of fact-checking and the understandability of verdicts.
The generated real-time explanation can also be used for educational purposes, helping users understand critical thinking processes and promoting public engagement.

\begin{table}[t]
\small
\centering
\begin{tabular}{lccc}
\hline
Split & SUPPORTED & REFUTED & NEI \\
\hline
Training & 80,035 & 29,775 & 35,639 \\
Development & 6,666 & 6,666 & 6,666 \\
Test & 6,666 & 6,666 & 6,666 \\ 
\hline
\end{tabular}
\caption{Statistics on the FEVER dataset.}
\label{tab: FEVER Statistics}
\end{table}

\section{Experiments}

\subsection{Settings} 

\paragraph{Datasets.} We evaluate our OE-Fact system on the Fact Extraction and VERification (FEVER) dataset~\citep{thorne2018fever}. 
As the largest and most popular dataset in automated fact verification, the FEVER dataset comprises claims constructed from the 2017 dump of Wikipedia, containing 185,445 claims and a corpus size of over 5,000,000 articles. 
Detailed statistic on the FEVER dataset is shown in Table~\ref{tab: FEVER Statistics}.

Since FEVER's test set is a blind test set without label annotations, we evaluate our fact-checking system on FEVER's development set across closed-domain and open-domain scenarios.
For the closed domain setting, the task is to verify the claim against predicted evidence from relevant Wikipedia page(s). 
For the open domain setting, the task is to verify the authenticity of claims based on evidence retrieved from open websites.

\paragraph{Baselines.} For our OE-Fact system, we use the large language model Meta AI 2 (Llama 2)\footnote{\url{https://github.com/facebookresearch/llama}}~\citep{touvron2023llama} as our foundation model.
We chose Llama 2 for two reasons: First, it is entirely open-source, allowing all individuals full access to the model for any purpose. 
Second, this language model does not require API access and thus does not create security issues for developers, providing further research and development possibilities.
We compare our OE-Fact system with the BEVERS\footnote{\url{https://github.com/mitchelldehaven/bevers}}~\cite{dehaven2023bevers} system, which is a standard three-module baseline system that carefully tunes modules for document retrieval, sentence selection, and claim classification and achieves SoTA performance on the FEVER dataset.

\paragraph{Evaluation.} We use accuracy (A{\small cc}) as well as precision (P), recall (R), and F1 scores across different labels to evaluate the proportion of correctly predicted claim labels, ignoring the evidence of the prediction.

\begin{table*}[t]
\small
\centering
\resizebox{\linewidth}{!}{
\begin{tabular}{lccccccccccc}
\hline
\multirow{2}{*}{Setting} & \multirow{2}{*}{Model} & \multirow{2}{*}{Accuracy (A)} & \multicolumn{3}{c}{SUPPORTED} & \multicolumn{3}{c}{REFUTED} & \multicolumn{3}{c}{NEI} \\
\cline{4-12}
~ & ~ & ~ & P & R & F1 & P & R & F1 & P & R & F1 \\
\hline
\multirow{2}{*}{Closed-domain} & BEVERS~\citep{dehaven2023bevers} & 33.91 & 34.1 & 36.7 & 25.0 & 33.2 & 33.6 & 22.3 & 34.5 & 31.4 & 21.7 \\
~ & OE-Fact & 49.75 & 58.6 & 56.9 & 57.7 & 46.2 & 78.4 & 58.1 & 42.2 & 13.9 & 11.7 \\
\hdashline
\multirow{2}{*}{Open-domain} & BEVERS~\citep{dehaven2023bevers} & 34.37 & 34.0 & 100.0 & 50.7 & 100.0 & 1.4 & 2.8 & 100.0 & 0.5 & 0.9 \\
~ & OE-Fact & 54.20 & 59.9 & 72.4 & 65.6 & 55.9 & 59.8 & 57.8 & 42.1 & 30.2 & 25.4 \\
\hline
\end{tabular}
}
\caption{Overall results on the FEVER development set.}
\label{tab:Overall Results}
\end{table*}

\begin{figure}
    \centering
    \includegraphics[height=4cm,width=8cm]{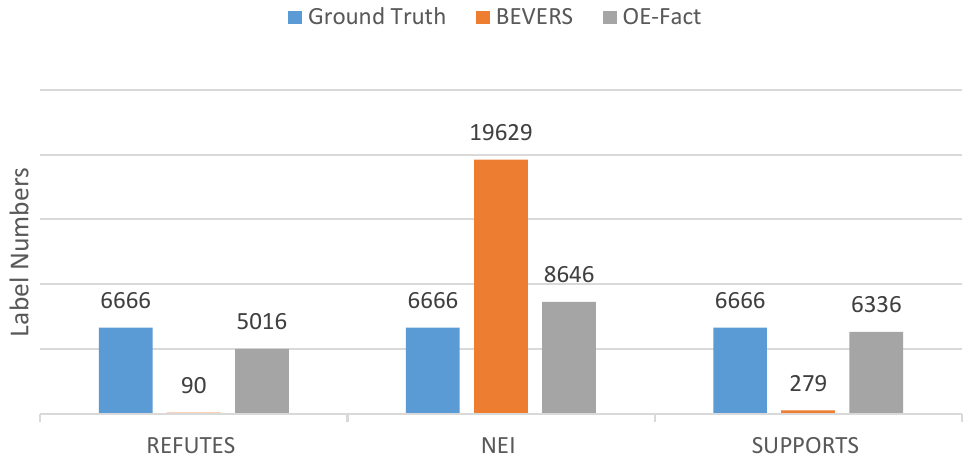}
    \caption{Label distribution comparison}
    \label{fig:label distribution}
\end{figure}

\begin{table*}[t]
\small
\centering
\begin{tabular}{lcccc}
\hline
Model & Accuracy (A) & SUPPORTED & REFUTED & NEI \\
\hline
LLM-3-module (OE-Fact) & \textbf{49.75} & \textbf{57.7} & 58.1 & 11.7 \\
LLM-2-module & 49.51 & 56.3 & \textbf{58.6} & \textbf{11.9} \\
LLM-1-module & 46.46 & 52.8 & 57.1 & 0.8 \\
\hdashline
BEVERS~\citep{dehaven2023bevers} & 33.91 & 25.0 & 22.3 & 21.7 \\
\hline
\end{tabular}
\caption{Results of LLM-based three-module framework on open-domain fact-checking. }
\label{tab:three-module}
\end{table*}

\subsection{Results}

The overall results are presented in Table~\ref{tab:Overall Results}.
Firstly, we compare our OE-Fact system with the BEVERS~\citep{dehaven2023bevers} system on the general FEVER fact-checking task in the closed-domain setting.
We rely on predicted evidence from the Wikipedia database to validate claims in this database-centric scenario.
Comparing the results obtained by OE-Fact and BEVERS (first two rows), we find that our OE-Fact system significantly improves the baseline system in terms of accuracy to 49.75, and all F1 scores in three labels are effectively improved.
This result intuitively demonstrates the effectiveness of our OE-Fact system in general fact-checking tasks.

Secondly, we compare OE-Fact and BEVERS in the open-domain setting.
Under this setting, we simulate real-world claim checking on the FEVER dataset, verifying the claim's authenticity based on the evidence retrieved from open websites. 
The results in Table~\ref{tab:Overall Results} show that our OE-Fact system significantly outperforms the benchmark BEVERS system and achieves 54.20\% on the accuracy metric.
In particular, comparing the overall results within closed- and open-domain settings, both the baseline BEVERS and OE-Fact perform better with sufficient real-time evidence.
However, from the label distribution in Figure~\ref{fig:label distribution}, we observed that the label distribution in the three categories generated by BEVERS is highly uneven, showing the limitation of traditional fact-checking models in real-world claim checking.
In contrast, our OE-Fact system performs more robustly and balanced in all three fact-checking categories, highlighting the effectiveness of our proposed OE-Fact system in both closed- and open-domain settings.

\section{Analysis}

\subsection{LLM-based Three-module Framework}

In our proposed OE-Fact system, to gradually inspire and stimulate the reasoning and judgment capabilities of LLM, we employ a three-module pipeline with LLM to carry out the claim-related evidence selection module and the claim verification and interpretation module step by step.
To test the effectiveness of our LLM-based three-module framework, we compare it with LLM-based one-module and two-module frameworks.
In the LLM-based two-module framework, We eliminate the claim-related evidence selection module and utilize all retrieved evidence for LLM-based claim verification and interpretation.
In the LLM-based one-module framework, we directly prompt LLM to verify and explain decisions for claims:

\vspace{3ex}
\begin{tabular}{p{6.5cm}} 
\emph{Verify the truth of \colorbox{lightgray}{Claim} with label `True', `False', or `Uncertain', give evidence and explain why you get this conclusion. Your output is: \textcolor{gray}{Label}; \textcolor{gray}{Evidence}; \textcolor{gray}{Explanation}.} \\
\end{tabular}
\vspace{3ex}

We report results for three-, two- and one-module frameworks on the development set of FEVER under the closed-domain setting and we report the results of BEVERS baseline for reference, as shown in Table~\ref{tab:three-module}.

\begin{table*}[t]
\small
\centering
\begin{tabular}{lp{12cm}}
\hline
Claim & Mass fish death in Malaysia caused by algal bloom, not Fukushima wastewater. \\
\hline 
\multirow{8}{*}{Verdict\&Explanation}  & Label: Uncertain. Explanation: The provided evidence 1, 2, 3, and 4 all suggest that mass fish death in Malaysia may be caused by algal blooms rather than Fukushima wastewater. However, the lack of understanding about the linkages between climate change and water quality impacts (evidence5) makes it uncertain whether this is indeed the case. Additionally, there are reports of bioluminescent algal blooms in other locations (evidence6-8), which could also contribute to mass fish deaths but are not necessarily related to Fukushima wastewater. Therefore, based on the available evidence we cannot conclude with certainty whether Mass fish death in Malaysia was caused by algal bloom or Fukushima wastewater. \\
\hline
\multirow{10}{*}{Evidence} & 1. When scientific evidence shows that a particular microorganism or its toxins can cause foodborne illness, the FDA may consider that microorganism to be capable ... \\
~ & “This is leading to growing instances of algal blooms, some of which are toxic, making it unsafe for humans and animals alike. \\
~ & Mar 16, 2023 ... Red tide effects on animals. The neurotoxins produced by harmful algae blooms can cause massive fish kills, with dead fish washing up on shores ... \\
~ & ... algal blooms. Linkages between climate change and water quality impacts are not well understood, however. Several factors explain this lack of understanding ... \\
~ & Apr 22, 2021 ... brevis blooms are the best studied of Florida HABs and include acute exposure impacts such as significant dies-offs of fish, marine mammals, ...\\
\hline
\end{tabular}
\caption{A real-world claim-checking example with filtered evidence, label verdict, and real-time fact-checking decision explanation. }
\label{tab:Example}
\end{table*}

It shows that all LLM-based fact-checking frameworks outperform the traditional three-module BEVERS.
Moreover, the LLM-based one-module framework, which verifies claims and generates explanations based on LLM's knowledge, obtains the lowest performance, showing limitation of the knowledge within our used LLM.
In contrast, both LLM-based two- and three-module fact-checking systems that retrieve sufficient evidence from the Wikipedia database, can significantly improve the performance.
Furthermore, by comparing the results of two- and three-module frameworks, we find that utilizing claim-related evidence filtered by the LLM-based evidence selection module for claim verification boost our OE-Fact system to the highest accuracy.
In general, experimental results demonstrate the effectiveness of our LLM-based claims-related evidence selection module.

\subsection{Decision Explanation}

We present an example of real-world claim checking from our OE-Fact system in Table~\ref{tab:Example}.
It contains filtered evidence, accurate verdicts, and real-time decision explanations to show the causal interpretability of the OE-Fact system.
As shown in the table, the generated explanation first enumerates which evidence clearly supports the label and then analyzes how the evidence supports or refutes the labels, thereby dialectically explaining the verdict.
This concise and coherent explanation ensures transparency about the accuracy of the verdict.

\section{Conclusion and Future Work}

This paper proposes an Open-domain Explainable Fact-checking (OE-Fact) system on the open-domain real-word claim-checking task.
With claim-related evidence retrieved from open websites and LLM-based claim-relevant evidence selection, our OE-Fact system effectively improves the reliability of claim verdicts and generates concise real-time fact-checking decision explanations.
It is worth noting that this work, for the first time, highlights the effectiveness of LLM in open-domain fact-checking and fills the gap in real-world claim verification scenarios.

In future work, we plan to further improve the reliability of claim verification and evaluate the logical coherence of real-time verdict explanations.



\bibliography{anthology,custom}
\bibliographystyle{acl_natbib}




\end{document}